\documentclass[11pt]{article}
\usepackage{nodalida2025}
\usepackage{times}
\usepackage{url}
\usepackage{booktabs}
\usepackage{latexsym}
\usepackage{microtype}
\usepackage{adjustbox}
\usepackage[hidelinks]{hyperref}
\usepackage{enumitem}
\setlist{noitemsep, topsep=0pt}
\usepackage[T1]{fontenc}
\aclfinalcopy 

\title{Train More Parameters But Mind Their Placement: \\ Insights into Language Adaptation with PEFT}

\author{Jenny Kunz \\
  Dept. of Computer and Information Science \\
  Linköping University \\
  {\tt jenny.kunz@liu.se} \\}

\date{}

\begin{document}
\maketitle
\begin{abstract}
    Smaller LLMs still face significant challenges even in medium-resourced languages, particularly when it comes to language-specific knowledge -- a problem not easily resolved with machine-translated data. 
    In this case study on Icelandic, we aim to enhance the generation performance of an LLM by specialising it using unstructured text corpora. A key focus is on preventing interference with the models’ capabilities of handling longer context during this adaptation. Through ablation studies using various parameter-efficient fine-tuning (PEFT) methods and setups, we find that increasing the number of trainable parameters leads to better and more robust language adaptation. LoRAs placed in the feed-forward layers and bottleneck adapters show promising results with sufficient parameters, while prefix tuning and (IA)$^3$ are not suitable. Although improvements are consistent in 0-shot summarisation, some adapted models struggle with longer context lengths, an issue that can be mitigated by adapting only the final layers. 
\end{abstract}

\section{Introduction} 

LLMs have strong multilingual capabilities and top the leaderboards even for less-represented languages \citep{nielsen2024encodervsdecodercomparative}. However, smaller LLMs still struggle with these languages, hampering fast and resource-efficient inference. 
Instruction tuning on machine-translated data can improve performance compared to English-only tuning \citep{muennighoff-etal-2023-crosslingual, chen-etal-2024-monolingual} but models still fall short when evaluated on native benchmarks, likely due to missing language-specific knowledge \citep{chen2024gooddatamultilingualinstruction}. 
While collecting large amounts of native instruction-tuning data could address this issue, this can be costly or infeasible. This makes techniques for adapting models using unstructured text data valuable. 

In this paper, we perform ablations with parameter-efficient fine-tuning (PEFT) methods for language adaptation with unstructured text data \textit{after} instruction alignment. This diverges from the standard setup for fine-tuning a model: Unlike typical fine-tuning, where the adaptation data closely matches the expected output format, the data we use is closer to the expected output in language but likely further from the target task format. Therefore, the setup risks interference with the original instruction-tuning objectives, possibly leading to \textit{catastrophic forgetting} \citep{mccloskey:catastrophic}. In addition, hardware constraints made us choose a maximum context length smaller than the one used in pre-training, risking further performance degradation. 

Therefore, we aim to identify setups that do not interfere with previously learned abilities. 
We attempt to avoid catastrophic forgetting with PEFT methods that leave the majority of or all model parameters unchanged: LoRA \citep{hu2022lora}, IA$^3$ \citep{liu2022fewshot}, bottleneck adapters \citep{pmlr-v97-houlsby19a} and prefix tuning \citep{li-liang-2021-prefix}.  
We experiment with the number of learnable parameters, the placement of LoRA matrices in different Transformer modules and layers, as well as the training corpus used for adaptation. 

We use the smallest instruction-tuned LLaMA 3.2 model \citep{dubey2024llama3herdmodels} with 1B parameters and adapt it to Icelandic, evaluating performance on text summarisation. Our findings are that: 
\begin{itemize}[leftmargin=*]
    \item LoRA and bottleneck adapters show improvements especially in 0-shot settings, though simply adding target-language task demonstrations also improves the performance substantially. 
    \item A higher number of trainable parameters is better.
    \item LoRAs in the feed-forward layers are the best-performing setup, followed by bottleneck adapters. LoRA in the attention layers works less well, particularly considering the number of trainable parameters. We therefore conclude that feed-forward modules are the most promising target in language adaptation. 
    \item Prefix tuning hurts the model's capabilities. 
    \item Some setups with few trainable parameters negatively impact 5-shot performance, possibly due to smaller context lengths at adaptation time compared to pre-training time. This can be resolved by restricting adapter placement to the top layers.
\end{itemize}

\section{Experimental Setup}

\subsection{Models} 
We use \textit{Llama-3.2-1B-Instruct}, the newest and smallest Llama model at the time of writing, with 1B, 16 layers, and a hidden size of 2048. This model has been tuned with instruction fine-tuning \citep{wei2022finetuned} and reinforcement learning with human feedback \citep{rlhf}\footnote{Ablations with the base model \textit{Llama-3.2-1B} showed inferior performance with and without adaptation. }. 

\subsection{Adaptation Data} 
Our main dataset for adaptation is the Icelandic portion of CC100 \citep{conneau-etal-2020-unsupervised} that has been processed with CCNet filtering \citep{wenzek-etal-2020-ccnet} to increase data quality. We randomly select 250,000 text chunks, with a maximum length of 1,024 tokens, resulting in 12.5M tokens. This data was likely seen during pre-training, i.e., the model is not exposed to new data but \textit{primed} towards Icelandic. 
As web-crawled corpora are reportedly of lower quality for smaller languages \citep{kreutzer, artetxe-etal-2022-corpus}, we perform ablations with the curated Icelandic Gigaword Corpus (IGC) \citep{steingrimsson-etal-2018-risamalheild, barkarson-etal-2022-evolving}, using sections from its subsets \textit{Books}, \textit{Wiki}, \textit{Social}, and \textit{Journals}. Even here we use 250,000 chunks, resulting in 12M tokens. As the \textit{Social} subset is by far the largest and we aim to have a large portion of highly curated text, we undersample it by using only 10\%, resulting in a dataset composition of 9\% \textit{Books}, 17\% \textit{Wiki}, 22\% \textit{Journals}, and 52\% \textit{Social}.
    
\subsection{Adaptation Methods and Setups}

The code, prompt generator and adapters used for the experiments in this paper can be found at \url{github.com/jekunz/peft-la}. We use the Transformers \citep{wolf-etal-2020-transformers} and Adapters \citep{poth-etal-2023-adapters} libraries, a learning rate of 5e-5, a linear learning rate scheduler, and a batch size of 4.\footnote{As the learning rate and scheduler are crucial in continued pre-training \citep{ibrahim2024simple}, we also tested 1e-5 and 1e-4 and a cosine scheduler but did not observe large differences.} All adapters are trained with a causal language modeling objective. We test the following methods and setups: 

\vspace{2pt}
\noindent \textbf{LoRA} is a widespread adaptation technique for generative LLMs. In the most common setup, it adds low-rank decomposition matrices to the model's self-attention modules and trains only those. The matrices can be merged into the weights, removing the inference overhead. 
For LoRA in the attention module, we test ranks 1024, 256, 128, 32 and 8 and apply LoRA to the query and value matrices, which is reportedly the most stable setup \citep{fomenko2024notelora}. 
We also test LoRA in the feed-forward module and place LoRAs in all matrices using ranks 256, 128, 64, 32 and 8. 
For both module setups and all ranks, we use use $\alpha = 2r$. 
\vspace{2pt}

\noindent \textbf{IA$^3$} is the most parameter-efficient among the methods tested. It multiplies activations in the model's attention (key and value) and feed-forward matrices with learned vectors, adding hardly any overhead. 
\vspace{2pt} 

\noindent \textbf{Bottleneck adapters} add smaller intermediate layers with a down- and up-projection in between the model's layers. While popular for encoder model, bottleneck adapters are less common for generative LLMs as they increase the number of parameters and depth even during inference. 
We train Houlsby adapters with reduction factors of 64, 16 and 4. 
\vspace{2pt} 

\noindent \textbf{Prefix tuning} prepends a sequence of learnable prefix vectors to the input sequences, allowing the model to attend to the prefix vectors when generating the subsequent tokens. As the vectors add to the sequence length, even prefix tuning slows down inference. 
We use a prefix length of 30 tokens. 

\subsection{Evaluation}

To assess generative performance, we evaluate abstractive text summarisation with the RÚV Radio News (RRN) dataset \citep{sverrisson-einarsson-2023-abstractive} in the \textit{main} $\rightarrow$ \textit{intro} setup, i.e., generating the introduction from the main body of the article. We filter out articles missing one of these fields.

We evaluate the summaries using BERTScore \citep{bertscore} (base model: \textit{bert-base-multilingual-uncased}) to measure the representational similarity between the output and the reference, and ROUGE-L \citep{lin-2004-rouge} for surface overlap, based on the longest common subsequence. 

The models are evaluated in 0-shot, 1-shot and 5-shot setups with minimal prompts in Icelandic\footnote{We also tested English instructions, which led to slightly worse results, except for the \textit{no adapters} model, where English instruction slightly improved the 0-shot performance. } that instruct the model to summarise the article in one paragraph and include markers for the start of both the article and the summary. 

\section{Results and Discussion}

%

\subsection{PEFT Methods}
\label{sec:results_peft}

\begin{table}[h]
\centering
\adjustbox{max width=\columnwidth}{%
\begin{tabular}{lccc}\toprule
& \multicolumn{1}{c}{0-shot} & \multicolumn{1}{c}{1-shot} & \multicolumn{1}{c}{5-shot}\\\midrule
No Adapter  & 53.37 / 04.09 & 64.68 / 10.26 & 64.01 / 11.37 \\\midrule
LoRA-qv-1024    & 63.61 / 08.57 & 66.53 / 11.70  & 65.50 / 12.06 \\
LoRA-qv-256     & 63.27 / 08.32 & 65.55 / 11.05 & 62.97 / 10.56 \\
LoRA-qv-128     & 62.55 / 07.63 & 64.51 / 10.78 & 62.23 / 10.54 \\ 
LoRA-qv-32      & 61.06 / 06.62 & 62.68 / 08.98 & 55.42 / 05.60 \\
LoRA-qv-8       & 60.45 / 05.21 & 61.53 / 08.23 & 56.62 / 06.42 \\\midrule
LoRA-ff-256     & \textbf{65.60 / 09.72} & \textbf{69.06 / 13.89} & \textbf{69.10 / 15.48} \\
LoRA-ff-128     & 64.67 / 08.87 & \textbf{69.10 / 13.86} &  68.36 / 14.55\\ 
LoRA-ff-64      & 63.72 / 07.86 & 67.72 / 12.60 & 67.46 / 13.65 \\
LoRA-ff-32      & 62.94 / 07.19 & 67.61 / 12.18 & 67.42 / 13.76  \\
LoRA-ff-8       & 61.69 / 06.36 & 64.85 / 10.39 & 62.66 / 10.09\\\midrule
(IA)$^3$        & 56.70 / 04.56 & 64.07 / 09.47 & 61.74 / 10.37 \\\midrule
Bottlen.\--4    & 63.78 / 08.15 & 66.75 / 11.74 & 66.74 / 13.21 \\
Bottlen.\--16   & 63.33 / 08.38 & 67.77 / 13.11 & 65.80 / 12.36 \\
Bottlen.\--64   & 60.66 / 05.16 & 64.79 / 09.96 & 61.32 / 08.59 \\\midrule
Prefix      & 55.84 / 02.02 & 54.56 / 01.73 & 49.86 / 00.67 \\\bottomrule
\end{tabular}}
    \caption{Comparing adaptation methods. BERTScore F1 / ROUGE-L.}
    \label{fig:peft_methods}
\end{table}

As shown in Table \ref{fig:peft_methods}, language adaptation consistently improves 0-shot summarisation scores. However, for 1-shot and 5-shot setups, the results are more mixed, and in some setups decrease compared to the baseline without adaptation. That the 1-shot setup without adaptation already shows comparable performance to many adaptation setups implies that in-context learning, where possible, can be an alternative to language adaptation for this model. 

\begin{figure*}
    \centering
    \includegraphics[width=1.00\linewidth]{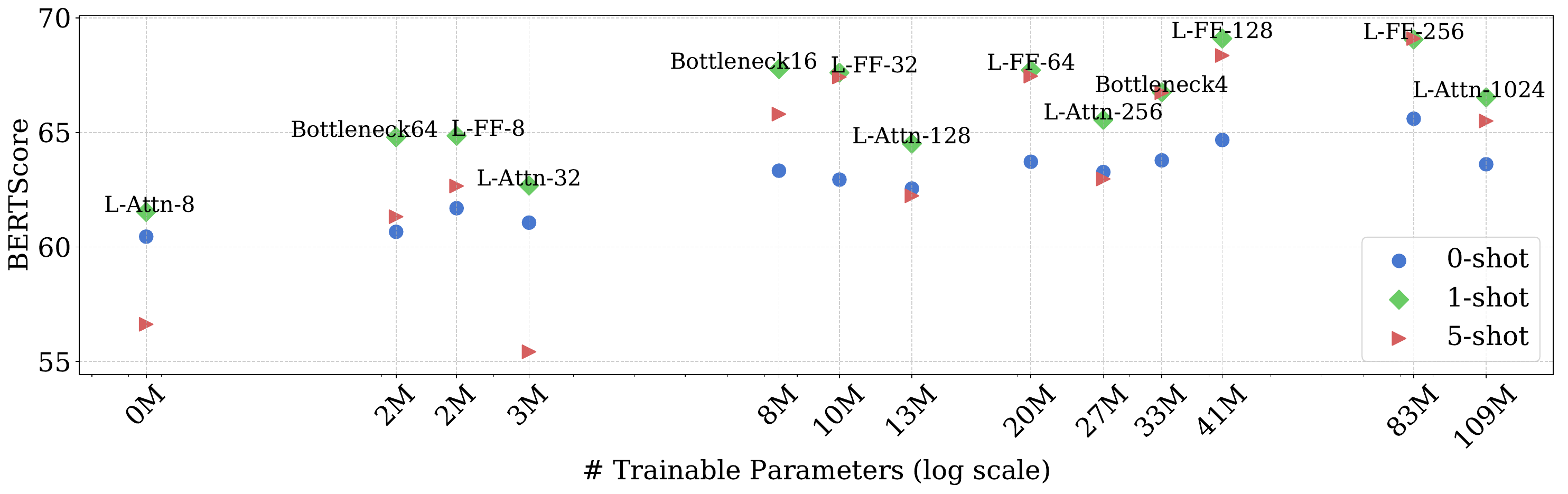}
    \caption{Number of trainable parameters plotted against BERTScores. Prefix tuning (34M parameters) and (IA)$^3$ (49K parameters) are excluded. }
    \label{fig:plot}
\end{figure*}

The best-performing method are LoRAs in the feed-forward layers. 
Even bottleneck adapters with a reduction factor of 16 or 4 consistently increase scores, although there is a noticeable difference in performance to feed-forward LoRA. As illustrated in Figure \ref{fig:plot}, feed-forward LoRA also results in the highest BERTScores relative to the number of parameters added, followed by bottleneck adapters. LoRA in the attention matrices requires substantially more parameters to reach a comparable performance. These results show that the placement of the PEFT modules in the Transformer architecture plays a crucial role even if the number of trainable parameters is the same. 

Some setups interfere with the model's ability to operate on longer inputs as the performance especially in the 5-shot setup decreases. We hypothesise this is a result of limiting the context length to 1,024 tokens during the adaptation process. LoRA in the attention module is the most heavily affected setup, suggesting that the effectiveness of self-attention when processing longer contexts is harmed.

We observe that performance improves as the LoRA rank increases or the bottleneck reduction factor decreases, indicating that sufficient learning capacity is necessary for better results in language adaptation. This is in line with the underwhelming performance of (IA)$^3$, which introduces the fewest parameters. Designed as an alternative to in-context learning for task adaptation, (IA)$^3$ does not transfer well to language adaptation. 

Prefix tuning with textual data decreases the performance substantially for the 1- and 5-shot setups. We assume that as prefixes have a direct impact on the generation, prefixes that diverge from the expected output format harm the model's abilities to match the latter. For this reason, prefix-tuning an instruction-tuned model on unlabelled text does not work, whereas prefix-tuning on specific tasks like summarisation, or instruction tuning in general, works well as shown by \citet{zhang2024llamaadapter}. 

\subsection{Ablation 1: LoRA Modules}

\begin{table}[h]
\centering
\adjustbox{max width=\columnwidth}{%
\begin{tabular}{lccc}\toprule
& \multicolumn{1}{c}{0-shot} & \multicolumn{1}{c}{1-shot} & \multicolumn{1}{c}{5-shot}\\\midrule
q,v         & 63.27 / 08.32 & 65.55 / 11.05 & 62.97 / 10.56 \\ 
ff          & 65.60 / 09.72 & 69.06 / 13.89 & 69.10 / 15.48 \\ 
ff + q,v    & 65.44 / 09.61 & 68.44 / 13.14 & 68.89 / 15.17 \\ 
\bottomrule
\end{tabular}}
    \caption{Comparing LoRA module placement. BERTScore F1 / ROUGE-L; LoRA rank 256}
    \label{fig:modules}
\end{table}
We have a closer look at the module placement of LoRAs and compare LoRA in the self-attention module, LoRA in the feed-forward module, and LoRA both in the self-attention and the feed-forward module. 

In the results given in Table \ref{fig:modules}, we see that for the same rank, LoRA in the feed-forward module is better than in the attention module. Moreover, it is slightly better than LoRA in both the attention and the feed-forward modules. We find this surprising given that the latter option has the most trainable parameters and conclude that having LoRA even in the attention matrices is at best unnecessary. 

\subsection{Ablation 2: Layer Exclusion}
\label{sec:layers}

\begin{table}[h]
\centering
\adjustbox{max width=\columnwidth}{%
\begin{tabular}{lccc}\toprule
& \multicolumn{1}{c}{0-shot} & \multicolumn{1}{c}{1-shot} & \multicolumn{1}{c}{5-shot}\\\midrule
No Adapter  & 53.37 / 04.09 & \textbf{64.68 / 10.26} & 64.01 / 11.37 \\
All Layers & \textbf{61.06 / 06.62} & 62.68 / 08.98 & 55.42 / 05.60 \\
All but last 2  & 59.39 / 04.83 & 60.26 / 07.37 & 57.73 / 06.51 \\
All but last 4  & 59.89 / 04.93 & 62.37 / 08.70 & 58.29 / 07.02 \\
Only last 2  & 59.64 / 03.64 & 63.55 / 08.37 & \textbf{65.40 / 12.42} \\
Only last 4  & 58.78 / 04.56 & 62.20 / 08.22 & 61.94 / 10.57 \\\bottomrule
\end{tabular}}
    \caption{Layer Exclusion experiments. BERTScore F1 / ROUGE-L;  Self-attention (qv) LoRA rank 32.} 
    \label{fig:layer_exclusion}
\end{table}
%

Fine-tuning primarily affects the final layers of a model \citep{merchant-etal-2020-happens, mosbach-etal-2020-interplay, zhou-srikumar-2022-closer}. We explore two strategies focusing on these layers: (1) \textit{excluding} the final layers during adaptation to preserve the instruction-tuning capabilities while focusing on general language learning, which is likely stored in earlier layers, and (2) adapting \textit{only} the final layers, as this may be sufficient and could maintain the model's robustness with respect to the limited context length used in our adaptation process (a key issue highlighted in Section \ref{sec:results_peft}). 

We test the two hypotheses using self-attention LoRA with rank 32 as this configuration shows strong 0-shot performance but suffers in the 5-shot setup. 
The results in Table \ref{fig:layer_exclusion} show that the first hypothesis does not hold; excluding the last layers does not improve the performance and, in some cases, degrades it. The second hypothesis, however, appears plausible: restricting LoRA modules to the last two layers yields the best 5-shot results among all setups in Table \ref{fig:layer_exclusion}, outperforming the baseline without adaptation. However, this comes at the expense of a slight decrease in 0-shot performance. 
We are hopeful that these insights can guide us in developing customised methods for language adaptation. 

\subsection{Ablation 3: Training Corpora}

\begin{table}[h]
\centering
\adjustbox{max width=\columnwidth}{%
\begin{tabular}{lccc}\toprule
& \multicolumn{1}{c}{0-shot} & \multicolumn{1}{c}{1-shot} & \multicolumn{1}{c}{5-shot}\\\midrule
CCNet & 63.27 / 08.32 & 65.55 / 11.05 & 62.97 / 10.56 \\
IGC &  60.80 / 05.75 & 61.02 / 06.48 & 58.31 / 06.17 \\\midrule
CCNet & 65.60 / 09.72 & 69.06 / 13.89 & 69.10 / 15.48 \\
IGC & 63.66 / 08.10 & 66.19 / 10.46 & 66.37 / 12.00 \\\midrule
CCNet & 63.78 / 08.15 & 66.75 / 11.74 & 66.74 / 13.21 \\
IGC & 61.39 / 05.58 & 64.95 / 09.79 &  65.24 / 11.54\\\bottomrule
\end{tabular}}
    \caption{Comparing text corpora for adaptation. BERTScore F1 / ROUGE-L; LoRA-qv-256 (above), LoRA-ff-256 (middle) and bottleneck reduction factor 4 (below).}
    \label{fig:corpora}
\end{table}

In Table \ref{fig:corpora}, we do not observe a benefit of training on the IGC; on the contrary, the performance is consistently lower. While this is in line with previous research \citep{artetxe-etal-2022-corpus, van-noord-etal-2024-language}, note that we do not test on any task where high-quality generation is important but on text summarisation, which can rely on copying chunks of text. We also note that CCNet is probably more diverse, and that different mixes from the IGC may lead to different results. We therefore believe that it is worthwhile to continue testing on curated data. 

\subsection{Future Work}

In order to test whether our findings generalise, we plan to extend our approach to other languages, larger models and adapters trained on more data, and to explore the effect of training on longer contexts. 
Based on our experiments on the placement and training of adapters in Section \ref{sec:layers}, we hope to find a sweet spot for language adaptation where no relevant information is overwritten but generation performance is improved. Inspiration could be taken from methods that automatically detect, and assign more parameters to, layers of particular importance \citep{zhang2023adaptive, yao2024layerwiseimportancemattersmemory}. 

A common approach to mitigate interference is episodic memories -- mixing in examples from previous tasks \citep{chaudhry2019tinyepisodicmemoriescontinual}, in our case, instruction-tuning data. This has shown promise in other works \citep{jiang-etal-2024-instruction, parmar2024reusedontretrainrecipe}, making it worthwhile to incorporate.  

One challenge in evaluating language adaptation methods is that automatic metrics for generative performance provide limited and potentially misleading insights. While running extensive human evaluations for all ablations in this paper is impractical, a human study of model outputs for the most promising setups, across a diverse set of prompts, should be included in future evaluations. 

\section{Related Work}

\citet{razumovskaia2024analyzingadaptinglargelanguage} find that LoRA language adaptation with unstructured text data improves the linguistic quality of generated texts in human ratings but usefulness and performance on a (translated) natural language inference benchmark remain low. Their study indicates that benchmark evaluation could underestimate the usefulness of language adaptation in chat and generation setups. 

Work on testing other PEFT architectures than LoRA for language adaptation of LLMs has been sparse. While bottleneck-style language adapters trained on text corpora are a common setup for cross-lingual transfer with encoder models \citep{pfeiffer-etal-2020-mad, he-etal-2021-effectiveness, faisal-anastasopoulos-2022-phylogeny}, they have been largely overlooked for generative models, likely due to the inference overhead that can be avoided with LoRA, as the latter works equally well for task fine-tuning. Our experiments show that similar findings hold for language adapters: Bottleneck adapters perform well but there are LoRA setups that reach the same performance or are better while avoiding the overhead.

Recent language adaptation works have focused on target-language instruction fine-tuning, often with machine-translated data \citep{muennighoff-etal-2023-crosslingual, chen-etal-2024-monolingual, holmstrom-doostmohammadi-2023-making}. In cross-lingual transfer, multilingual instruction tuning has shown promise, particularly for generative tasks \citep{kew2023turning} and for larger models \citep{chen-etal-2024-monolingual}. However, models trained on machine-translated data may perform well on translated evaluation sets but struggle on native benchmarks \citep{chen2024gooddatamultilingualinstruction}. 


\section{Conclusion}
\label{sec:conclusion}

We tested a range of PEFT methods for language adaptation using unstructured text corpora, finding that LoRA in the feed-forward modules yielded the most promising results, followed by bottleneck adapters. LoRA in the attention modules performed less well, was less robust to larger context lengths and needed more parameters for a comparable performance. Combining LoRAs in both the attention and feed-forward modules did not improve over feed-forward LoRAs only, and may even lead to slightly decreased performance. 
Prefix tuning and (IA)$^3$ were not suitable at all. 

Our results show that across architectures, more trainable parameters lead to better scores, showing, perhaps unsurprisingly, that sufficient learning capacity is crucial for language adaptation. 

Some adaptation setups led to a decline in performance as contexts get longer; possibly a result of restricted context lengths during adaptation. However, this issue can be mitigated by training only the last layers. Notably, we did not observe any positive effects from using higher-quality pre-training data sourced from narrower domains. 

Moving forward, with a higher resource investment, we see the potential that more training data, possibly with instruction data in the mix, and longer context lengths improve the performance further. However, to truly assess the potential of these methods, we need more diverse, language-native evaluation data, as well as fine-grained human evaluations that assess various aspects of generated language quality and content.

\section*{Limitations}

The meaningfulness of automated text summarisation metrics when using news text summaries as references has been questioned and is highly dependent on the dataset \citep{zhang-etal-2024-benchmarking}. While our search for effective setups yielded conclusive results with BERTScore and ROUGE-L, 
moving forward, it will be crucial to incorporate human evaluations and more diverse tasks to accurately assess performance across a broader and better-interpretable range of criteria. 

As we have discussed in Section \ref{sec:conclusion}, we see a critical need for more language-native evaluation data, in particular datasets that incorporate significant language-specific knowledge \citep{chen2024gooddatamultilingualinstruction}. Testing on a limited set of language-native tasks most of which are classification tasks, or on machine-translated data, may give a limited picture of the effect of language adaptation. 

Due to computational constraints, we were unable to include larger models or more than one language in this study. As a result, it remains unclear whether our findings apply to other languages, especially those that are typologically more different from or closer to English. 


\section*{Acknowledgments}

I thank my colleagues Kevin Glocker, Kätriin Kukk, Julian Schlenker, Marcel Bollmann, Noah-Manuel Michael and Romina Oji for valuable discussions at all stages of this project and feedback on earlier drafts, and the anonymous reviewers for their constructive feedback and insightful suggestions. 

This work was partially supported by the Wallenberg AI, Autonomous Systems and Software Program (WASP) funded by the Knut and Alice Wallenberg Foundation. It is associated with TrustLLM funded by Horizon Europe GA 101135671. 
The computations were enabled by the Berzelius resource provided by the Knut and Alice Wallenberg Foundation at the National Supercomputer Centre and by the National Academic Infrastructure for Supercomputing in Sweden (NAISS), partially funded by the Swedish Research Council through grant agreement no. 2022-06725. 

\bibliographystyle{acl_natbib}
\bibliography{nodalida2025}

\begin{thebibliography}{42}
\expandafter\ifx\csname natexlab\endcsname\relax\def\natexlab#1{#1}\fi

\bibitem[{Artetxe et~al.(2022)Artetxe, Aldabe, Agerri, Perez-de Vi{\~n}aspre, and Soroa}]{artetxe-etal-2022-corpus}
Mikel Artetxe, Itziar Aldabe, Rodrigo Agerri, Olatz Perez-de Vi{\~n}aspre, and Aitor Soroa. 2022.
\newblock \href {https://doi.org/10.18653/v1/2022.emnlp-main.499} {Does corpus quality really matter for low-resource languages?}
\newblock In \emph{Proceedings of the 2022 Conference on Empirical Methods in Natural Language Processing}, pages 7383--7390, Abu Dhabi, United Arab Emirates. Association for Computational Linguistics.

\bibitem[{Barkarson et~al.(2022)Barkarson, Steingr{\'\i}msson, and Hafsteinsd{\'o}ttir}]{barkarson-etal-2022-evolving}
Starka{\dh}ur Barkarson, Stein{\th}{\'o}r Steingr{\'\i}msson, and Hildur Hafsteinsd{\'o}ttir. 2022.
\newblock \href {https://aclanthology.org/2022.lrec-1.254} {Evolving large text corpora: Four versions of the {I}celandic {G}igaword corpus}.
\newblock In \emph{Proceedings of the Thirteenth Language Resources and Evaluation Conference}, pages 2371--2381, Marseille, France. European Language Resources Association.

\bibitem[{Chaudhry et~al.(2019)Chaudhry, Rohrbach, Elhoseiny, Ajanthan, Dokania, Torr, and Ranzato}]{chaudhry2019tinyepisodicmemoriescontinual}
Arslan Chaudhry, Marcus Rohrbach, Mohamed Elhoseiny, Thalaiyasingam Ajanthan, Puneet~K. Dokania, Philip H.~S. Torr, and Marc'Aurelio Ranzato. 2019.
\newblock \href {http://arxiv.org/abs/1902.10486} {On tiny episodic memories in continual learning}.

\bibitem[{Chen et~al.(2024{\natexlab{a}})Chen, Ji, Bogoychev, Kutuzov, Haddow, and Heafield}]{chen-etal-2024-monolingual}
Pinzhen Chen, Shaoxiong Ji, Nikolay Bogoychev, Andrey Kutuzov, Barry Haddow, and Kenneth Heafield. 2024{\natexlab{a}}.
\newblock \href {https://aclanthology.org/2024.findings-eacl.90} {Monolingual or multilingual instruction tuning: Which makes a better alpaca}.
\newblock In \emph{Findings of the Association for Computational Linguistics: EACL 2024}, pages 1347--1356, St. Julian{'}s, Malta. Association for Computational Linguistics.

\bibitem[{Chen et~al.(2024{\natexlab{b}})Chen, Yu, Guo, and Haddow}]{chen2024gooddatamultilingualinstruction}
Pinzhen Chen, Simon Yu, Zhicheng Guo, and Barry Haddow. 2024{\natexlab{b}}.
\newblock \href {http://arxiv.org/abs/2406.12822} {Is it good data for multilingual instruction tuning or just bad multilingual evaluation for large language models?}

\bibitem[{Conneau et~al.(2020)Conneau, Khandelwal, Goyal, Chaudhary, Wenzek, Guzm{\'a}n, Grave, Ott, Zettlemoyer, and Stoyanov}]{conneau-etal-2020-unsupervised}
Alexis Conneau, Kartikay Khandelwal, Naman Goyal, Vishrav Chaudhary, Guillaume Wenzek, Francisco Guzm{\'a}n, Edouard Grave, Myle Ott, Luke Zettlemoyer, and Veselin Stoyanov. 2020.
\newblock \href {https://doi.org/10.18653/v1/2020.acl-main.747} {Unsupervised cross-lingual representation learning at scale}.
\newblock In \emph{Proceedings of the 58th Annual Meeting of the Association for Computational Linguistics}, pages 8440--8451, Online. Association for Computational Linguistics.

\bibitem[{Faisal and Anastasopoulos(2022)}]{faisal-anastasopoulos-2022-phylogeny}
Fahim Faisal and Antonios Anastasopoulos. 2022.
\newblock \href {https://aclanthology.org/2022.aacl-main.34} {Phylogeny-inspired adaptation of multilingual models to new languages}.
\newblock In \emph{Proceedings of the 2nd Conference of the Asia-Pacific Chapter of the Association for Computational Linguistics and the 12th International Joint Conference on Natural Language Processing (Volume 1: Long Papers)}, pages 434--452, Online only. Association for Computational Linguistics.

\bibitem[{Fomenko et~al.(2024)Fomenko, Yu, Lee, Hsieh, and Chen}]{fomenko2024notelora}
Vlad Fomenko, Han Yu, Jongho Lee, Stanley Hsieh, and Weizhu Chen. 2024.
\newblock \href {http://arxiv.org/abs/2404.05086} {A note on lora}.

\bibitem[{He et~al.(2021)He, Liu, Ye, Tan, Ding, Cheng, Low, Bing, and Si}]{he-etal-2021-effectiveness}
Ruidan He, Linlin Liu, Hai Ye, Qingyu Tan, Bosheng Ding, Liying Cheng, Jiawei Low, Lidong Bing, and Luo Si. 2021.
\newblock \href {https://doi.org/10.18653/v1/2021.acl-long.172} {On the effectiveness of adapter-based tuning for pretrained language model adaptation}.
\newblock In \emph{Proceedings of the 59th Annual Meeting of the Association for Computational Linguistics and the 11th International Joint Conference on Natural Language Processing (Volume 1: Long Papers)}, pages 2208--2222, Online. Association for Computational Linguistics.

\bibitem[{Holmstr{\"o}m and Doostmohammadi(2023)}]{holmstrom-doostmohammadi-2023-making}
Oskar Holmstr{\"o}m and Ehsan Doostmohammadi. 2023.
\newblock \href {https://aclanthology.org/2023.nodalida-1.62} {Making instruction finetuning accessible to non-{E}nglish languages: A case study on {S}wedish models}.
\newblock In \emph{Proceedings of the 24th Nordic Conference on Computational Linguistics (NoDaLiDa)}, pages 634--642, T{\'o}rshavn, Faroe Islands. University of Tartu Library.

\bibitem[{Houlsby et~al.(2019)Houlsby, Giurgiu, Jastrzebski, Morrone, De~Laroussilhe, Gesmundo, Attariyan, and Gelly}]{pmlr-v97-houlsby19a}
Neil Houlsby, Andrei Giurgiu, Stanislaw Jastrzebski, Bruna Morrone, Quentin De~Laroussilhe, Andrea Gesmundo, Mona Attariyan, and Sylvain Gelly. 2019.
\newblock \href {https://proceedings.mlr.press/v97/houlsby19a.html} {Parameter-efficient transfer learning for {NLP}}.
\newblock In \emph{Proceedings of the 36th International Conference on Machine Learning}, volume~97 of \emph{Proceedings of Machine Learning Research}, pages 2790--2799. PMLR.

\bibitem[{Hu et~al.(2022)Hu, yelong shen, Wallis, Allen-Zhu, Li, Wang, Wang, and Chen}]{hu2022lora}
Edward~J Hu, yelong shen, Phillip Wallis, Zeyuan Allen-Zhu, Yuanzhi Li, Shean Wang, Lu~Wang, and Weizhu Chen. 2022.
\newblock \href {https://openreview.net/forum?id=nZeVKeeFYf9} {Lo{RA}: Low-rank adaptation of large language models}.
\newblock In \emph{International Conference on Learning Representations}.

\bibitem[{Ibrahim et~al.(2024)Ibrahim, Th{\'e}rien, Gupta, Richter, Anthony, Belilovsky, Lesort, and Rish}]{ibrahim2024simple}
Adam Ibrahim, Benjamin Th{\'e}rien, Kshitij Gupta, Mats~Leon Richter, Quentin~Gregory Anthony, Eugene Belilovsky, Timoth{\'e}e Lesort, and Irina Rish. 2024.
\newblock \href {https://openreview.net/forum?id=DimPeeCxKO} {Simple and scalable strategies to continually pre-train large language models}.
\newblock \emph{Transactions on Machine Learning Research}.

\bibitem[{Jiang et~al.(2024)Jiang, Sun, Shi, Rodriguez, Zhou, Neubig, Lin, Yih, and Iyer}]{jiang-etal-2024-instruction}
Zhengbao Jiang, Zhiqing Sun, Weijia Shi, Pedro Rodriguez, Chunting Zhou, Graham Neubig, Xi~Lin, Wen-tau Yih, and Srini Iyer. 2024.
\newblock \href {https://doi.org/10.18653/v1/2024.acl-long.296} {Instruction-tuned language models are better knowledge learners}.
\newblock In \emph{Proceedings of the 62nd Annual Meeting of the Association for Computational Linguistics (Volume 1: Long Papers)}, pages 5421--5434, Bangkok, Thailand. Association for Computational Linguistics.

\bibitem[{Kew et~al.(2023)Kew, Schottmann, and Sennrich}]{kew2023turning}
Tannon Kew, Florian Schottmann, and Rico Sennrich. 2023.
\newblock \href {http://arxiv.org/abs/2312.12683} {Turning english-centric llms into polyglots: How much multilinguality is needed?}

\bibitem[{Kreutzer et~al.(2022)Kreutzer, Caswell, Wang, Wahab, van Esch, Ulzii-Orshikh, Tapo, Subramani, Sokolov, Sikasote, Setyawan, Sarin, Samb, Sagot, Rivera, Rios, Papadimitriou, Osei, Suarez, Orife, Ogueji, Rubungo, Nguyen, Müller, Müller, Muhammad, Muhammad, Mnyakeni, Mirzakhalov, Matangira, Leong, Lawson, Kudugunta, Jernite, Jenny, Firat, Dossou, Dlamini, de~Silva, Çabuk Ballı, Biderman, Battisti, Baruwa, Bapna, Baljekar, Azime, Awokoya, Ataman, Ahia, Ahia, Agrawal, and Adeyemi}]{kreutzer}
Julia Kreutzer, Isaac Caswell, Lisa Wang, Ahsan Wahab, Daan van Esch, Nasanbayar Ulzii-Orshikh, Allahsera Tapo, Nishant Subramani, Artem Sokolov, Claytone Sikasote, Monang Setyawan, Supheakmungkol Sarin, Sokhar Samb, Benoît Sagot, Clara Rivera, Annette Rios, Isabel Papadimitriou, Salomey Osei, Pedro~Ortiz Suarez, Iroro Orife, Kelechi Ogueji, Andre~Niyongabo Rubungo, Toan~Q. Nguyen, Mathias Müller, André Müller, Shamsuddeen~Hassan Muhammad, Nanda Muhammad, Ayanda Mnyakeni, Jamshidbek Mirzakhalov, Tapiwanashe Matangira, Colin Leong, Nze Lawson, Sneha Kudugunta, Yacine Jernite, Mathias Jenny, Orhan Firat, Bonaventure F.~P. Dossou, Sakhile Dlamini, Nisansa de~Silva, Sakine Çabuk Ballı, Stella Biderman, Alessia Battisti, Ahmed Baruwa, Ankur Bapna, Pallavi Baljekar, Israel~Abebe Azime, Ayodele Awokoya, Duygu Ataman, Orevaoghene Ahia, Oghenefego Ahia, Sweta Agrawal, and Mofetoluwa Adeyemi. 2022.
\newblock \href {https://doi.org/10.1162/tacl_a_00447} {{Quality at a Glance: An Audit of Web-Crawled Multilingual Datasets}}.
\newblock \emph{Transactions of the Association for Computational Linguistics}, 10:50--72.

\bibitem[{Li and Liang(2021)}]{li-liang-2021-prefix}
Xiang~Lisa Li and Percy Liang. 2021.
\newblock \href {https://doi.org/10.18653/v1/2021.acl-long.353} {Prefix-tuning: Optimizing continuous prompts for generation}.
\newblock In \emph{Proceedings of the 59th Annual Meeting of the Association for Computational Linguistics and the 11th International Joint Conference on Natural Language Processing (Volume 1: Long Papers)}, pages 4582--4597, Online. Association for Computational Linguistics.

\bibitem[{Lin(2004)}]{lin-2004-rouge}
Chin-Yew Lin. 2004.
\newblock \href {https://aclanthology.org/W04-1013} {{ROUGE}: A package for automatic evaluation of summaries}.
\newblock In \emph{Text Summarization Branches Out}, pages 74--81, Barcelona, Spain. Association for Computational Linguistics.

\bibitem[{Liu et~al.(2022)Liu, Tam, Mohammed, Mohta, Huang, Bansal, and Raffel}]{liu2022fewshot}
Haokun Liu, Derek Tam, Muqeeth Mohammed, Jay Mohta, Tenghao Huang, Mohit Bansal, and Colin Raffel. 2022.
\newblock \href {https://openreview.net/forum?id=rBCvMG-JsPd} {Few-shot parameter-efficient fine-tuning is better and cheaper than in-context learning}.
\newblock In \emph{Advances in Neural Information Processing Systems}.

\bibitem[{LlamaTeam(2024)}]{dubey2024llama3herdmodels}
LlamaTeam. 2024.
\newblock \href {http://arxiv.org/abs/2407.21783} {The llama 3 herd of models}.

\bibitem[{Mccloskey and Cohen(1989)}]{mccloskey:catastrophic}
Michael Mccloskey and Neil~J. Cohen. 1989.
\newblock Catastrophic interference in connectionist networks: {T}he sequential learning problem.
\newblock \emph{The Psychology of Learning and Motivation}, 24:104--169.

\bibitem[{Merchant et~al.(2020)Merchant, Rahimtoroghi, Pavlick, and Tenney}]{merchant-etal-2020-happens}
Amil Merchant, Elahe Rahimtoroghi, Ellie Pavlick, and Ian Tenney. 2020.
\newblock \href {https://doi.org/10.18653/v1/2020.blackboxnlp-1.4} {What happens to {BERT} embeddings during fine-tuning?}
\newblock In \emph{Proceedings of the Third BlackboxNLP Workshop on Analyzing and Interpreting Neural Networks for NLP}, pages 33--44, Online. Association for Computational Linguistics.

\bibitem[{Mosbach et~al.(2020)Mosbach, Khokhlova, Hedderich, and Klakow}]{mosbach-etal-2020-interplay}
Marius Mosbach, Anna Khokhlova, Michael~A. Hedderich, and Dietrich Klakow. 2020.
\newblock \href {https://doi.org/10.18653/v1/2020.blackboxnlp-1.7} {On the interplay between fine-tuning and sentence-level probing for linguistic knowledge in pre-trained transformers}.
\newblock In \emph{Proceedings of the Third BlackboxNLP Workshop on Analyzing and Interpreting Neural Networks for NLP}, pages 68--82, Online. Association for Computational Linguistics.

\bibitem[{Muennighoff et~al.(2023)Muennighoff, Wang, Sutawika, Roberts, Biderman, Le~Scao, Bari, Shen, Yong, Schoelkopf, Tang, Radev, Aji, Almubarak, Albanie, Alyafeai, Webson, Raff, and Raffel}]{muennighoff-etal-2023-crosslingual}
Niklas Muennighoff, Thomas Wang, Lintang Sutawika, Adam Roberts, Stella Biderman, Teven Le~Scao, M~Saiful Bari, Sheng Shen, Zheng~Xin Yong, Hailey Schoelkopf, Xiangru Tang, Dragomir Radev, Alham~Fikri Aji, Khalid Almubarak, Samuel Albanie, Zaid Alyafeai, Albert Webson, Edward Raff, and Colin Raffel. 2023.
\newblock \href {https://doi.org/10.18653/v1/2023.acl-long.891} {Crosslingual generalization through multitask finetuning}.
\newblock In \emph{Proceedings of the 61st Annual Meeting of the Association for Computational Linguistics (Volume 1: Long Papers)}, pages 15991--16111, Toronto, Canada. Association for Computational Linguistics.

\bibitem[{Nielsen et~al.(2024)Nielsen, Enevoldsen, and Schneider-Kamp}]{nielsen2024encodervsdecodercomparative}
Dan~Saattrup Nielsen, Kenneth Enevoldsen, and Peter Schneider-Kamp. 2024.
\newblock \href {http://arxiv.org/abs/2406.13469} {Encoder vs decoder: Comparative analysis of encoder and decoder language models on multilingual nlu tasks}.

\bibitem[{van Noord et~al.(2024)van Noord, Kuzman, Rupnik, Ljube{\v{s}}i{\'c}, Espl{\`a}-Gomis, Ram{\'\i}rez-S{\'a}nchez, and Toral}]{van-noord-etal-2024-language}
Rik van Noord, Taja Kuzman, Peter Rupnik, Nikola Ljube{\v{s}}i{\'c}, Miquel Espl{\`a}-Gomis, Gema Ram{\'\i}rez-S{\'a}nchez, and Antonio Toral. 2024.
\newblock \href {https://aclanthology.org/2024.lrec-main.465} {Do language models care about text quality? evaluating web-crawled corpora across 11 languages}.
\newblock In \emph{Proceedings of the 2024 Joint International Conference on Computational Linguistics, Language Resources and Evaluation (LREC-COLING 2024)}, pages 5221--5234, Torino, Italia. ELRA and ICCL.

\bibitem[{Ouyang et~al.(2024)Ouyang, Wu, Jiang, Almeida, Wainwright, Mishkin, Zhang, Agarwal, Slama, Ray, Schulman, Hilton, Kelton, Miller, Simens, Askell, Welinder, Christiano, Leike, and Lowe}]{rlhf}
Long Ouyang, Jeff Wu, Xu~Jiang, Diogo Almeida, Carroll~L. Wainwright, Pamela Mishkin, Chong Zhang, Sandhini Agarwal, Katarina Slama, Alex Ray, John Schulman, Jacob Hilton, Fraser Kelton, Luke Miller, Maddie Simens, Amanda Askell, Peter Welinder, Paul Christiano, Jan Leike, and Ryan Lowe. 2024.
\newblock Training language models to follow instructions with human feedback.
\newblock In \emph{Proceedings of the 36th International Conference on Neural Information Processing Systems}, NIPS '22, Red Hook, NY, USA. Curran Associates Inc.

\bibitem[{Parmar et~al.(2024)Parmar, Satheesh, Patwary, Shoeybi, and Catanzaro}]{parmar2024reusedontretrainrecipe}
Jupinder Parmar, Sanjev Satheesh, Mostofa Patwary, Mohammad Shoeybi, and Bryan Catanzaro. 2024.
\newblock \href {http://arxiv.org/abs/2407.07263} {Reuse, don't retrain: A recipe for continued pretraining of language models}.

\bibitem[{Pfeiffer et~al.(2020)Pfeiffer, Vuli{\'c}, Gurevych, and Ruder}]{pfeiffer-etal-2020-mad}
Jonas Pfeiffer, Ivan Vuli{\'c}, Iryna Gurevych, and Sebastian Ruder. 2020.
\newblock \href {https://doi.org/10.18653/v1/2020.emnlp-main.617} {{MAD-X}: {A}n {A}dapter-{B}ased {F}ramework for {M}ulti-{T}ask {C}ross-{L}ingual {T}ransfer}.
\newblock In \emph{Proceedings of the 2020 Conference on Empirical Methods in Natural Language Processing (EMNLP)}, pages 7654--7673, Online. Association for Computational Linguistics.

\bibitem[{Poth et~al.(2023)Poth, Sterz, Paul, Purkayastha, Engl{\"a}nder, Imhof, Vuli{\'c}, Ruder, Gurevych, and Pfeiffer}]{poth-etal-2023-adapters}
Clifton Poth, Hannah Sterz, Indraneil Paul, Sukannya Purkayastha, Leon Engl{\"a}nder, Timo Imhof, Ivan Vuli{\'c}, Sebastian Ruder, Iryna Gurevych, and Jonas Pfeiffer. 2023.
\newblock \href {https://aclanthology.org/2023.emnlp-demo.13} {Adapters: A unified library for parameter-efficient and modular transfer learning}.
\newblock In \emph{Proceedings of the 2023 Conference on Empirical Methods in Natural Language Processing: System Demonstrations}, pages 149--160, Singapore. Association for Computational Linguistics.

\bibitem[{Razumovskaia et~al.(2024)Razumovskaia, Vulić, and Korhonen}]{razumovskaia2024analyzingadaptinglargelanguage}
Evgeniia Razumovskaia, Ivan Vulić, and Anna Korhonen. 2024.
\newblock \href {http://arxiv.org/abs/2403.01929} {Analyzing and adapting large language models for few-shot multilingual nlu: Are we there yet?}

\bibitem[{Steingr{\'\i}msson et~al.(2018)Steingr{\'\i}msson, Helgad{\'o}ttir, R{\"o}gnvaldsson, Barkarson, and Gu{\dh}nason}]{steingrimsson-etal-2018-risamalheild}
Stein{\th}{\'o}r Steingr{\'\i}msson, Sigr{\'u}n Helgad{\'o}ttir, Eir{\'\i}kur R{\"o}gnvaldsson, Starka{\dh}ur Barkarson, and J{\'o}n Gu{\dh}nason. 2018.
\newblock \href {https://aclanthology.org/L18-1690} {{R}isam{\'a}lheild: A very large {I}celandic text corpus}.
\newblock In \emph{Proceedings of the Eleventh International Conference on Language Resources and Evaluation ({LREC} 2018)}, Miyazaki, Japan. European Language Resources Association (ELRA).

\bibitem[{Sverrisson and Einarsson(2023)}]{sverrisson-einarsson-2023-abstractive}
{\TH}{\'o}r Sverrisson and Hafsteinn Einarsson. 2023.
\newblock \href {https://aclanthology.org/2023.nodalida-1.3} {Abstractive text summarization for {I}celandic}.
\newblock In \emph{Proceedings of the 24th Nordic Conference on Computational Linguistics (NoDaLiDa)}, pages 17--31, T{\'o}rshavn, Faroe Islands. University of Tartu Library.

\bibitem[{Wei et~al.(2022)Wei, Bosma, Zhao, Guu, Yu, Lester, Du, Dai, and Le}]{wei2022finetuned}
Jason Wei, Maarten Bosma, Vincent Zhao, Kelvin Guu, Adams~Wei Yu, Brian Lester, Nan Du, Andrew~M. Dai, and Quoc~V Le. 2022.
\newblock \href {https://openreview.net/forum?id=gEZrGCozdqR} {Finetuned language models are zero-shot learners}.
\newblock In \emph{International Conference on Learning Representations}.

\bibitem[{Wenzek et~al.(2020)Wenzek, Lachaux, Conneau, Chaudhary, Guzm{\'a}n, Joulin, and Grave}]{wenzek-etal-2020-ccnet}
Guillaume Wenzek, Marie-Anne Lachaux, Alexis Conneau, Vishrav Chaudhary, Francisco Guzm{\'a}n, Armand Joulin, and Edouard Grave. 2020.
\newblock \href {https://aclanthology.org/2020.lrec-1.494} {{CCN}et: Extracting high quality monolingual datasets from web crawl data}.
\newblock In \emph{Proceedings of the Twelfth Language Resources and Evaluation Conference}, pages 4003--4012, Marseille, France. European Language Resources Association.

\bibitem[{Wolf et~al.(2020)Wolf, Debut, Sanh, Chaumond, Delangue, Moi, Cistac, Rault, Louf, Funtowicz, Davison, Shleifer, von Platen, Ma, Jernite, Plu, Xu, Le~Scao, Gugger, Drame, Lhoest, and Rush}]{wolf-etal-2020-transformers}
Thomas Wolf, Lysandre Debut, Victor Sanh, Julien Chaumond, Clement Delangue, Anthony Moi, Pierric Cistac, Tim Rault, Remi Louf, Morgan Funtowicz, Joe Davison, Sam Shleifer, Patrick von Platen, Clara Ma, Yacine Jernite, Julien Plu, Canwen Xu, Teven Le~Scao, Sylvain Gugger, Mariama Drame, Quentin Lhoest, and Alexander Rush. 2020.
\newblock \href {https://doi.org/10.18653/v1/2020.emnlp-demos.6} {Transformers: State-of-the-art natural language processing}.
\newblock In \emph{Proceedings of the 2020 Conference on Empirical Methods in Natural Language Processing: System Demonstrations}, pages 38--45, Online. Association for Computational Linguistics.

\bibitem[{Yao et~al.(2024)Yao, Gao, Li, Zhao, Wang, Wang, and Zhu}]{yao2024layerwiseimportancemattersmemory}
Kai Yao, Penlei Gao, Lichun Li, Yuan Zhao, Xiaofeng Wang, Wei Wang, and Jianke Zhu. 2024.
\newblock \href {http://arxiv.org/abs/2410.11772} {Layer-wise importance matters: Less memory for better performance in parameter-efficient fine-tuning of large language models}.

\bibitem[{Zhang et~al.(2023)Zhang, Chen, Bukharin, He, Cheng, Chen, and Zhao}]{zhang2023adaptive}
Qingru Zhang, Minshuo Chen, Alexander Bukharin, Pengcheng He, Yu~Cheng, Weizhu Chen, and Tuo Zhao. 2023.
\newblock \href {https://openreview.net/forum?id=lq62uWRJjiY} {Adaptive budget allocation for parameter-efficient fine-tuning}.
\newblock In \emph{The Eleventh International Conference on Learning Representations}.

\bibitem[{Zhang et~al.(2024{\natexlab{a}})Zhang, Han, Liu, Zhou, Lu, Qiao, Li, and Gao}]{zhang2024llamaadapter}
Renrui Zhang, Jiaming Han, Chris Liu, Aojun Zhou, Pan Lu, Yu~Qiao, Hongsheng Li, and Peng Gao. 2024{\natexlab{a}}.
\newblock \href {https://openreview.net/forum?id=d4UiXAHN2W} {{LL}a{MA}-adapter: Efficient fine-tuning of large language models with zero-initialized attention}.
\newblock In \emph{The Twelfth International Conference on Learning Representations}.

\bibitem[{Zhang et~al.(2020)Zhang, Kishore, Wu, Weinberger, and Artzi}]{bertscore}
Tianyi Zhang, Varsha Kishore, Felix Wu, Kilian~Q. Weinberger, and Yoav Artzi. 2020.
\newblock \href {https://openreview.net/forum?id=SkeHuCVFDr} {Bertscore: Evaluating text generation with {BERT}}.
\newblock In \emph{8th International Conference on Learning Representations, {ICLR} 2020, Addis Ababa, Ethiopia, April 26-30, 2020}. OpenReview.net.

\bibitem[{Zhang et~al.(2024{\natexlab{b}})Zhang, Ladhak, Durmus, Liang, McKeown, and Hashimoto}]{zhang-etal-2024-benchmarking}
Tianyi Zhang, Faisal Ladhak, Esin Durmus, Percy Liang, Kathleen McKeown, and Tatsunori~B. Hashimoto. 2024{\natexlab{b}}.
\newblock \href {https://doi.org/10.1162/tacl_a_00632} {Benchmarking large language models for news summarization}.
\newblock \emph{Transactions of the Association for Computational Linguistics}, 12:39--57.

\bibitem[{Zhou and Srikumar(2022)}]{zhou-srikumar-2022-closer}
Yichu Zhou and Vivek Srikumar. 2022.
\newblock \href {https://doi.org/10.18653/v1/2022.acl-long.75} {A closer look at how fine-tuning changes {BERT}}.
\newblock In \emph{Proceedings of the 60th Annual Meeting of the Association for Computational Linguistics (Volume 1: Long Papers)}, pages 1046--1061, Dublin, Ireland. Association for Computational Linguistics.

\end{thebibliography}

\end{document}